\documentclass{article}
\usepackage{subcaption}

\usepackage[final]{neurips_2025}

\usepackage[breakable]{tcolorbox} 
\tcbuselibrary{skins}
\tcbuselibrary{raster}
\usepackage{graphicx}
\usepackage{booktabs}
\usepackage{colortbl}
\usepackage{dblfloatfix}

\usepackage{adjustbox}
\usepackage[utf8]{inputenc} 
\usepackage[T1]{fontenc}    
\usepackage{hyperref}       
\usepackage{url}            
\usepackage{amsfonts}       
\usepackage{nicefrac}       
\usepackage{microtype}      
\usepackage{enumitem}
\usepackage{amsmath}

\usepackage{float}
\usepackage{graphicx}      
\usepackage{booktabs}      
\usepackage{array}         
\usepackage{colortbl}      
\definecolor{navyblue}{rgb}{0.0, 0.0, 0.5}
\definecolor{oai-green-200}{RGB}{204, 255, 204}
\definecolor{oai-green-400}{RGB}{153, 255, 153}
\definecolor{oai-green-600}{RGB}{102, 255, 102}
\definecolor{oai-gray-300}{RGB}{200, 200, 200}
\definecolor{oai-gray-600}{RGB}{150, 150, 150}
\definecolor{orange!10}{RGB}{255, 230, 204}
\definecolor{yellow!10}{RGB}{255, 255, 204}
\usepackage{caption}       
\usepackage{multirow}      
\usepackage{rotating}      
\usepackage{pifont}    
\usepackage{arydshln}
\usepackage{xspace}
\usepackage{fontawesome}

\usepackage{lineno}

\definecolor{prompt-color}{HTML}{B0B0B0}
\definecolor{chosen-color}{HTML}{808080}
\definecolor{rejected-color}{HTML}{808080}
\definecolor{lightblue}{RGB}{80,120,130}
\definecolor{mintgreen}{RGB}{60,180,60}
\definecolor{lavender}{RGB}{120,100,180}

\newcommand{\github}{\raisebox{-1.5pt}{\includegraphics[height=1.05em]{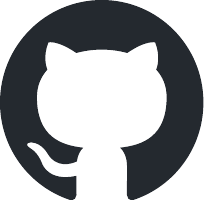}}\xspace}
\newcommand{\huggingface}{\raisebox{-1.5pt}{\includegraphics[height=1.05em]{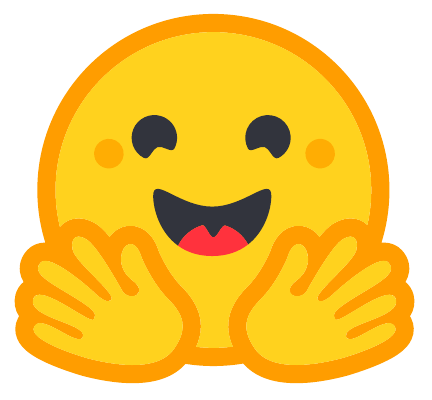}}\xspace}

\title{MIRAGE: A Multi-modal Benchmark for Spatial Perception, Reasoning, and Intelligence}

\author{
  Chonghan Liu$^{1}$\thanks{Equal contribution.} \\ 
  \and  
  Haoran Wang$^{2}$\footnotemark[1]\\
  \and
  Felix Henry$^{1}$ \\
  \and
  Pu Miao$^{3}$\\
  \and
  Yajie Zhang$^{1}$\\
  \and
  Yu Zhao$^{4}$\thanks{Equal advising.} \quad Peiran Wu$^{5}$\footnotemark[2]\\[1ex]
  \begin{minipage}{\textwidth}
    \centering
    $^{1}$Independent Researcher \quad
    $^{2}$Tsinghua University \quad
    $^{3}$shopee \\
    $^{4}$Alibaba International Digital Commerce \quad
    $^{5}$University of Bristol\\[0.5ex]
    {\github\ \href{https://github.com/khazic/Mirage}{\text{Evaluation Code}} \quad
    \huggingface\ \href{https://huggingface.co/datasets/Mmoment/Mirage_Multimodal_Benchmark}{\text{Mirage Bench}}}
  \end{minipage}
}

\begin{document}

\maketitle

\begin{abstract}
Spatial perception and reasoning are core components of human cognition, encompassing object recognition, spatial relational understanding, and dynamic reasoning. Despite progress in computer vision, existing benchmarks reveal significant gaps in models' abilities to accurately recognize object attributes and reason about spatial relationships, both essential for dynamic reasoning. To address these limitations, we propose MIRAGE, a multi-modal benchmark designed to evaluate models' capabilities in Counting (object attribute recognition), Relation (spatial relational reasoning), and Counting with Relation. Through diverse and complex scenarios requiring fine-grained recognition and reasoning, MIRAGE highlights critical limitations in state-of-the-art models, underscoring the need for improved representations and reasoning frameworks. By targeting these foundational abilities, MIRAGE provides a pathway toward spatiotemporal reasoning in future research.
\end{abstract}
\section{Introduction}

\begin{figure*}[htp]
    \vspace{-1em}
    \centering
    \includegraphics[width=\linewidth]{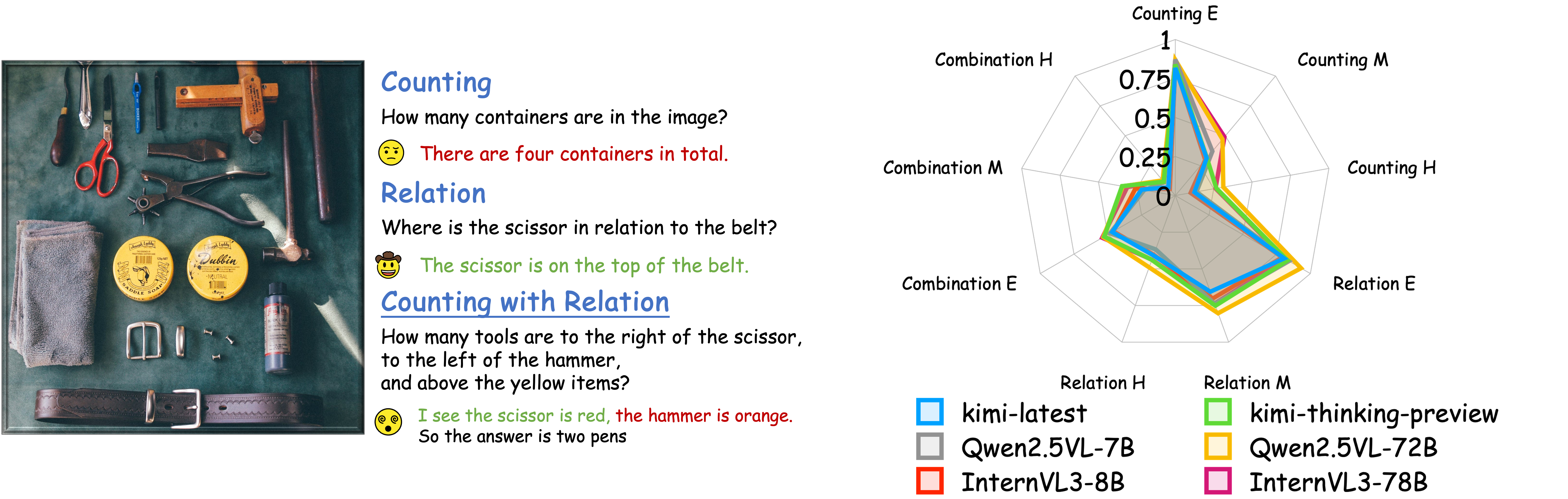}
    \vspace{-0.3em}
    \caption{
    \textit{Left:} Examples of our three task types—\textbf{Counting}, \textbf{Relation}, and \textbf{Counting with Relation}. Tasks increase in complexity as they require understanding object attributes, spatial relationships, and their composition. 
    \textit{Right:} Model performance across difficulty tiers (Easy, Medium, Hard) and task types. State-of-the-art models show consistent drops in the \textbf{Combination with Relation} setting, revealing weaknesses in compositional spatial reasoning.
    }
    \label{fig:teaser}
\end{figure*}

Cognition is a hierarchical process that underpins human perception of the visual world. At its foundation lies the ability to recognize objects—identifying entities and distinguishing them across varying conditions such as shape, size, color, or occlusion. Building on this, humans have developed spatial reasoning: understanding where objects are located relative to others, and how they interact within structured environments. Finally, humans acquire the capacity for dynamic reasoning—predicting how object configurations change over time and responding adaptively. This progression—from recognition, to spatial inference, to dynamic modeling—forms the basis of human intelligent behavior.

Consider a child playing with a ball on a table. To interpret this scene, one must first identify the relevant entities: what is the child, the ball, and the table as separate objects in the environment. Next, spatial relationships must be established: the ball is “on the table,” the child is “reaching for the ball,” and the table “supports” the ball. Finally, one may reason about future dynamics: the child may grasp the ball, or the ball may fall if nudged. Crucially, these levels of reasoning are compositional: the ability to reason about movement or change depends on correctly identifying objects and understanding how they relate to one another.

Modern vision-language models have made significant progress in object recognition, aided by large-scale pretraining and alignment. However, spatial relational reasoning—understanding where objects are in relation to each other and performing logic over those relationships—remains an open challenge due to imperfect image splitting strategy for training. This is especially evident in two foundational yet underexplored tasks: \textbf{Counting} (quantifying objects based on shared attributes), and \textbf{Relation} (understanding spatial positioning and reference). These tasks test not only visual perception, but also grounding, composition, and scene understanding.

Counting, while seemingly straightforward, often fails in real-world settings where objects vary in appearance or are partially occluded. VLMs may over-rely on statistical priors ("there are usually 2 chairs") rather than grounded instance reasoning. Likewise, relation tasks challenge models to localize objects with respect to spatial referents like “left of the cup” or “behind the child in red,” requiring both visual attention and symbolic spatial alignment. These reasoning demands escalate further in combinatorial tasks—e.g., “How many objects are to the left of the kettle and above the red container?”—which expose deeper integration failures.

To address these challenges, we introduce \textbf{MIRAGE}, a multi-modal benchmark that evaluates VLMs’ ability to perform object-centric reasoning and spatial composition. MIRAGE includes three task variants—\textbf{Counting}, \textbf{Relation}, and \textbf{Counting with Relation}—constructed over diverse images with grounded annotations and difficulty labels. Our design emphasizes the compositional nature of visual cognition, and targets the gap between surface-level recognition and relational understanding.

\textbf{Our contributions are summarized as follows:}
\begin{itemize}
    \item We introduce \textbf{MIRAGE}, a benchmark designed to evaluate object-centric and spatial reasoning through Counting, Relation, and Counting with Relation.
    \item We demonstrate that VLMs exhibit sharp drops on spatially-composed tasks, especially under occlusion, ambiguity, or referential complexity.
    \item We perform a series of diagnostic studies—including prompt tuning, spatial robustness tests, and error typology—that reveal the key challenges in grounded spatial reasoning.
\end{itemize}
\section{Related Works}

\subsection{Multimodal Large Language Models}
Multimodal large language models integrate vision and language modalities, enabling advanced capabilities in tasks like visual reasoning, captioning, and multimodal dialogue~\citep{alayrac2022flamingovisuallanguagemodel, dosovitskiy2021imageworth16x16words, li2022blipbootstrappinglanguageimagepretraining, radford2021learningtransferablevisualmodels}. Recent advancements in MLLMs like NaViT~\citep{dehghani2023patchnpacknavit}, Qwen-VL~\citep{bai2023qwenvlversatilevisionlanguagemodel, wang2024qwen2vlenhancingvisionlanguagemodels} and Kimi-VL~\citep{kimiteam2025kimivltechnicalreport} focus on improving architectural flexibility, such as dynamic resolution mechanisms and Mixture-of-Experts (MoE) designs. Concurrently, works like InternVL~\citep{chen2024fargpt4vclosinggap, chen2024internvlscalingvisionfoundation, zhu2025internvl3exploringadvancedtraining} and PixMo~\citep{deitke2024molmopixmoopenweights} highlight the importance of high-quality datasets and scalable training pipelines. Additionally, instruction-tuned models such as LLaVA~\citep{liu2023visualinstructiontuning} leverage synthetic multimodal data for enhanced generalization, while the MiniCPM family~\citep{hu2024minicpmunveilingpotentialsmall, yao2024minicpmvgpt4vlevelmllm} explores resource-efficient designs. These innovations collectively drive a shift toward efficient, flexible, and accessible MLLMs suitable for diverse real-world applications.

\subsection{Quantitative Object Understanding in Multimodal Models}
Quantitative object understanding, such as object enumeration and attribute quantification, remains a key challenge for multimodal large language models (MLLMs). While models excel in tasks like image captioning, they struggle with grounded enumeration, often hallucinating numbers or relying on statistical priors~\citep{xu2023lvlmehubcomprehensiveevaluationbenchmark, zhang2024goodcaptioningbadcounting}. These limitations are particularly evident in specialized domains such as Earth observation and remote sensing, where precise quantification is critical for applications like urban monitoring and disaster management~\citep{cai2025spatialbotprecisespatialunderstanding, danish2025geobenchvlmbenchmarkingvisionlanguagemodels}. Benchmarks like GEOBench-VLM~\citep{danish2025geobenchvlmbenchmarkingvisionlanguagemodels} and LVLM-eHub~\citep{xu2023lvlmehubcomprehensiveevaluationbenchmark} have highlighted these deficiencies, demonstrating that even state-of-the-art models fail to leverage clear object boundaries for accurate quantification. Addressing these gaps is crucial for deploying MLLMs in real-world scenarios requiring reliable object attribute cognition.

\subsection{Spatial and Spatial-Temporal Reasoning in Multimodal Models}
Spatial and spatial-temporal reasoning remain significant hurdles, particularly in understanding positional relationships and reasoning over depth, occlusion, and temporal dynamics~\citep{daxberger2025mmspatialexploring3dspatial, rajabi2024gsrbenchbenchmarkgroundedspatial}. Benchmarks like GSR-BENCH~\citep{rajabi2024gsrbenchbenchmarkgroundedspatial}, iVISPAR~\citep{mayer2025ivisparinteractivevisualspatial}, and MM-Spatial~\citep{daxberger2025mmspatialexploring3dspatial} reveal that while models perform better on 2D spatial tasks, they struggle with 3D and 4D contexts, such as spatial-temporal localization in egocentric videos~\citep{wu2025st, yang2024thinkingspacemultimodallarge}. Tasks like multi-step spatial reasoning and interactive planning~\citep{mayer2025ivisparinteractivevisualspatial} further expose the limitations of current architectures, which rely on pattern matching instead of robust representations. Recent efforts, such as STI-Bench~\citep{li2025stibenchmllmsreadyprecise}, emphasize precise spatial-temporal understanding, providing a pathway for human-like reasoning in dynamic environments.
\section{MIRAGE Benchmark}
\label{sec:construction}
\begin{figure*}[htp]
    \vspace{-1em}
    \centering
    \includegraphics[width=\linewidth]{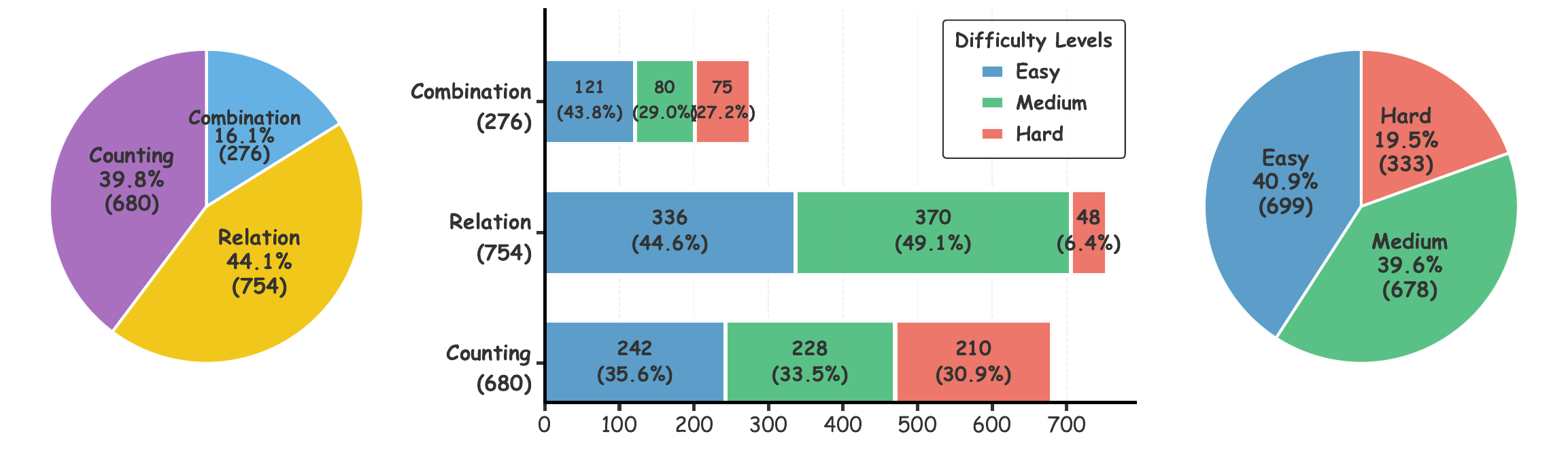}
    \vspace{-0.3em}
    \caption{
    \textbf{Dataset composition and difficulty breakdown in MIRAGE.} 
    \textit{Left:} Distribution of the three task types: \textbf{Counting} (39.8\%), \textbf{Relation} (44.1\%), and \textbf{Counting \& Relation} (16.1\%). 
    \textit{Middle:} Difficulty stratification (Easy, Medium, Hard) within each task type. \textit{Right:} Overall difficulty distribution across the entire dataset: 40.9\% Easy, 39.6\% Medium, and 19.5\% Hard.
    }
    \label{fig:construction}
\end{figure*}

\subsection{Task Definition}

MIRAGE evaluates two core aspects of visual reasoning: counting and relation. These tasks test a model’s ability to understand object attributes, infer spatial relationships, and combine these skills in complex scenes that simulate real-world complexities.

\textbf{Why Counting?}  
Counting measures a model's ability to answer questions like "How many objects of a specific type are in the scene?" This task challenges models to recognize objects across varying attributes, such as color, size, and texture. For example, counting apples requires identifying them whether they are red or green, big or small, fully visible or partially hidden. Additionally, counting demands precise localization to avoid errors like double-counting or missing occluded objects. Many scenes also include overlapping or partially occluded objects, making it necessary for models to reason about incomplete or ambiguous information. These challenges reflect the complexities of real-world scenarios and make counting an essential test of a model's generalization ability.

\textbf{Why Relation?}  
Relation reasoning evaluates a model’s understanding of spatial relationships between objects. For instance, determining if a ball is "on top of" a table or "under" a chair requires interpreting relative positions and contextual arrangements. Unlike counting, relation tasks emphasize interactions between objects rather than their individual features. Complex setups, such as "a book inside a bag on a table," test a model’s ability to reason about nested or hierarchical relationships. These tasks also introduce ambiguities due to overlapping or occluded objects, requiring the model to infer spatial configurations from limited visual cues. Such reasoning capabilities are critical for tasks like navigation, manipulation, and scene understanding in real-world environments.

\textbf{Task Composition: Counting with Relation}  
MIRAGE combines Counting and Relation tasks to evaluate a model's ability to integrate multiple reasoning skills. For instance, a task might ask, "How many balls are on the table?" Solving such queries requires simultaneous reasoning about object identity, quantity, and spatial configuration. These tasks introduce further complexity via diverse object appearances (e.g., color, shape, size, texture), as well as complex spatial arrangements like overlapping or nested objects. This integrated approach significantly raises task complexity, revealing gaps in current vision models' ability to generalize across diverse and ambiguous visual conditions. Our experiments show that such tasks expose critical limitations in state-of-the-art models, emphasizing the need for further advancements in visual reasoning.

By incorporating challenges such as diverse object features, occlusions, and intricate spatial setups, MIRAGE ensures a robust evaluation of a model’s visual reasoning capabilities, reflecting the complexities of real-world scenes and driving progress in static visual intelligence.

\begin{figure}[t!]
    \centering
    \includegraphics[width=\linewidth]{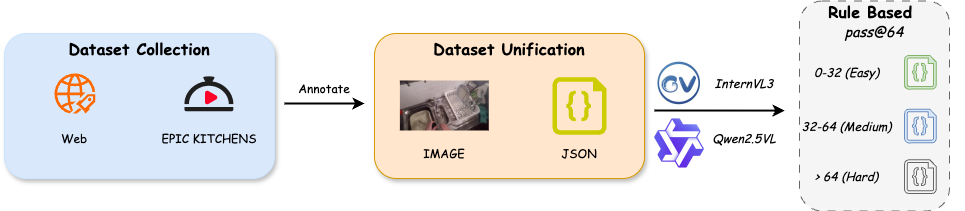}
    \vspace{0.2cm}
    \caption{
We collect images from both curated sources (e.g., EPIC-KITCHENS) and web-scale retrieval, followed by manual annotation of spatial reasoning tasks. Each image-question pair is unified into a standard format with aligned JSON labels. We then assign difficulty levels using a rule-based strategy: samples are evaluated by InternVL3 and Qwen2.5VL models with \texttt{pass@64}, and labeled as \textbf{Hard} (0–2 correct), \textbf{Medium} (2–16), or \textbf{Easy} ($>$16).
}
    \vspace{-0.4cm}
    \label{fig:pipeline}
\end{figure}

\subsection{Benchmark Construction}

To evaluate the reasoning capabilities of vision-language models across diverse visual tasks, we constructed a dataset specifically designed to challenge three core reasoning abilities: counting, relational reasoning, and their combination. The dataset consists of 1,710 questions, with 680 for counting, 754 for relational reasoning, and 276 for combination tasks. All questions were manually annotated and underwent a rigorous review process to ensure high-quality and consistent labels. 

\begin{figure}[t!]
    \centering
    \includegraphics[width=\linewidth]{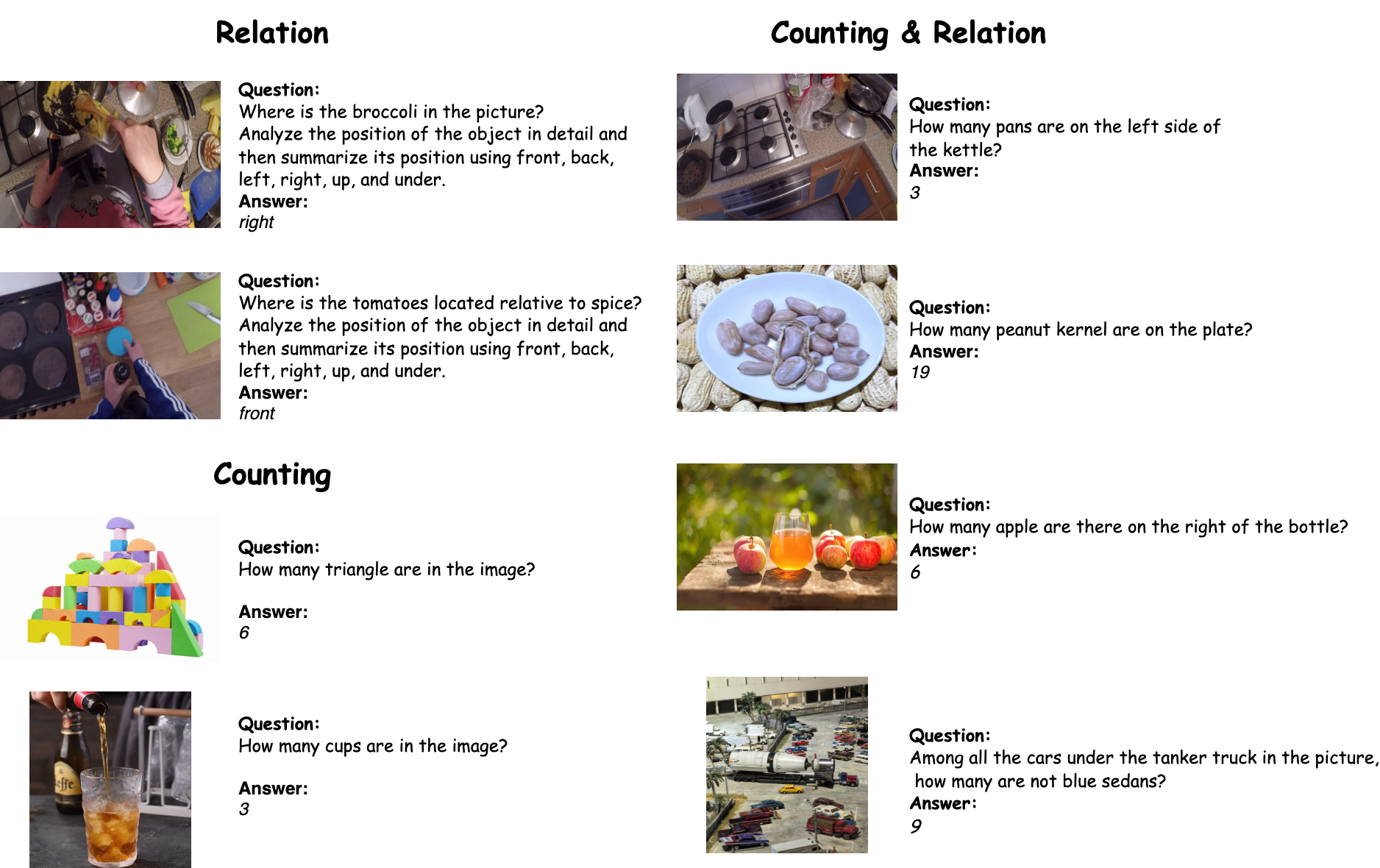}
    \vspace{0.2cm}
    \caption{We illustrate the three core task types in MIRAGE: \textbf{Counting} (top-left), which focuses on identifying and enumerating object instances; \textbf{Relation} (bottom-left), which involves locating objects using spatial references such as “left,” “right,” or “under”; and \textbf{Counting \& Relation} (right), which combines both reasoning types, requiring models to ground quantities within spatial constraints. Questions are designed to vary in complexity, visual context, and linguistic structure, highlighting the diverse challenges posed by the benchmark.
}
    \vspace{-0.4cm}
    \label{fig:main}
\end{figure}

\subsubsection{Visual Diversity and Reasoning Challenges}

MIRAGE draws from a rich mix of publicly available datasets, web sources, and original photography to maximize diversity in content, style, and context. The resulting images span egocentric perspectives, commercial imagery, artistic compositions, and everyday scenes. This visual variety is paired with a broad spectrum of reasoning challenges. Objects differ widely in color, shape, size, and texture; spatial relations range from simple pairs to deeply nested hierarchies; and many queries require the integration of both object-level and relational reasoning. Together, these factors create a benchmark that pushes models beyond surface-level recognition and exposes limitations in real-world generalization. A complete breakdown of data sources is provided in Appendix~\ref{app:data_source}.

\subsubsection{Tiny Subset and Difficulty Tiers}

To facilitate fast diagnosis and scalable evaluation, we provide two curated variants of MIRAGE: a \textbf{Tiny subset} and a \textbf{difficulty-tiered} full version. The Tiny subset consists of 50 representative questions spanning a range of difficulty. Specifically, it includes 10 examples correctly answered by both \texttt{Qwen2.5VL-3B} and \texttt{Qwen2.5VL-72B} (easy), 10 examples where only the stronger model succeeds (medium), and 30 examples where both models fail (hard). This compact version preserves task diversity while enabling lightweight evaluation for ablations or model iteration.

For the full benchmark, we further assign difficulty levels to each question based on model consensus. Using a \texttt{pass@64} metric evaluated over \texttt{InternVL-2.5-4B} and \texttt{Qwen2.5VL-3B}, we label a sample as \textbf{Hard} if both models succeed fewer than 2 times, \textbf{Medium} if between 2 and 16 completions are correct, and \textbf{Easy} if either model succeeds more than 16 times. This rule-based stratification reflects realistic model performance boundaries and allows detailed analysis of robustness across task types.

\section{Experiments}
\label{sec:exps}
We conduct a comprehensive and systematic set of experiments to evaluate the spatial and compositional reasoning capabilities of state-of-the-art vision-language models (VLMs) using MIRAGE. Beyond reporting aggregate accuracy, our goal is to uncover \emph{why} models succeed or fail under different forms of visual-linguistic stress.

To this end, we structure our analysis around three core questions:
\begin{enumerate}
\item Are performance bottlenecks primarily due to misunderstanding the task instructions, or to fundamental limitations in visual grounding?
\item How do current VLMs cope with real-world spatial challenges such as occlusion, crowding, or referential ambiguity?
\item Can prompting models to reason “step by step” improve grounded understanding, or does it risk introducing new forms of hallucination?
\end{enumerate}

We begin by benchmarking various proprietary and open-source models. Then, through targeted ablations and case studies, we diagnose the nature of model errors and evaluate whether structural prompt cues or reasoning scaffolds can mitigate them—or instead reveal deeper perceptual limitations.

\subsection{Main Results}

Across Table~\ref{tab:model_performance}, models perform best on Relation, moderately on Counting, and significantly worse on Counting with Relation, confirming the increased difficulty of reasoning about quantities under spatial constraints. Even large-scale models such as \texttt{Qwen2.5VL-72B} and \texttt{InternVL3-78B} show a $\sim$20-point drop in accuracy when moving from Relation to Combination, revealing fundamental limitations in compositional spatial reasoning.

\begin{table*}[h!]
\centering
\small
\renewcommand{\arraystretch}{1.2} 
\begin{adjustbox}{max width=\textwidth}
\begin{tabular}{@{}lcccccc@{}}
\toprule
\textbf{Model} & \textbf{Count (full)} & \textbf{Count (tiny)} & \textbf{Rel (full)} & \textbf{Rel (tiny)} & \textbf{Comb (full)} & \textbf{Comb (tiny)} \\ 
\midrule
\rowcolor[gray]{0.9}
\multicolumn{7}{c}{\textbf{Proprietary Models (Closed-source)}} \\ 
Kimi-latest & 43.85 & 36.00 & 71.33 & 46.00 & 28.77 & 32.00 \\ 
Kimi-thinking-preview & 49.82 & 38.00 & 82.61 & 54.00 & 34.17 & 36.00 \\ 
Claude-3-sonnet & - & 32.00 & - & 22.00 & - & 24.00 \\ 
Claude-3.5-sonnet & - & \textbf{48.00} & - & 58.00 & - & 50.00 \\ 
Claude-3-haiku & - & 28.00 & - & 4.00 & - & 18.00 \\ 
Gemini-2.0-flash-001 & - & \textbf{48.00} & - & \textbf{60.00} & - & \textbf{52.00} \\ 
GPT-4o-mini-0718 & - & 42.00 & - & 36.00 & - & 28.00 \\ 
GPT-4o-0513 & - & \textbf{48.00} & - & 54.00 & - & 36.00 \\ 
QwenVL-max & - & \textbf{48.00} & - & 42.00 & - & 44.00 \\ 
\rowcolor[gray]{0.9}
\multicolumn{7}{c}{\textbf{Open-source Models}} \\ 
QwenVL-2.5-3B & 38.33 & 30.00 & 74.47 & 52.00 & 23.83 & \textbf{40.00} \\ 
QwenVL-2.5-7B & 48.03 & 38.00 & 81.46 & 52.00 & 29.96 & 34.00 \\ 
QwenVL-2.5-72B & \textbf{56.62} & 40.00 & \textbf{85.31} & \textbf{58.00} & 36.94 & \textbf{40.00} \\ 
InternVL-3-8B & 44.24 & 38.00 & 75.32 & 40.00 & 29.24 & 38.00 \\ 
InternVL-3-78B & 55.15 & \textbf{46.00} & 82.60 & 56.00 & \textbf{36.10} & \textbf{40.00} \\ 
\bottomrule
\end{tabular}
\end{adjustbox}
\caption{
\textbf{Performance on the MIRAGE benchmark across three task types: Counting, Relation, and Counting with Relation, evaluated on both the full benchmark and the tiny diagnostic subset.} For closed-source models (e.g., Gemini, Claude, GPT-4o), we report results only on the tiny subset due to API cost constraints. As shown, performance trends on the tiny subset are consistent with those on the full benchmark, making it a reliable proxy for broader evaluation.
}
\label{tab:model_performance}
\end{table*}
As expected, scaling up model size improves accuracy: \texttt{Qwen2.5VL-3B} trails its 72B counterpart by over 13 points on full Counting (38.33\% $\rightarrow$ 56.62\%) and similarly on Combination (23.83\% $\rightarrow$ 36.94\%). However, the gap remains sizable even for frontier models, indicating that architectural improvements alone are insufficient.

For proprietary models, due to high API costs, we evaluate only on the tiny subset. Despite the smaller size, the relative performance ordering aligns closely with the full set, supporting its use as a trend-preserving proxy. For instance, \texttt{Gemini-2.0} and \texttt{Claude-3.5} both achieve over 50\% on Combination (tiny), outperforming most models, yet they still fall short of robust generalization, with many failure cases involving occlusion, ambiguous references, or compositional prompts.

Overall, these results highlight that while modern VLMs are competent at isolated spatial or counting tasks, integrating both under realistic visual ambiguity remains a major open challenge.

\subsection{How does VLMs perceive the world?}

\subsubsection{Prompt Modifications Improve Counting and Compositional Reasoning}

To investigate whether model errors stem from misunderstanding task instructions or from weaknesses in spatial reasoning, we modify the default prompt format for \texttt{InternVL3-8B} and evaluate performance across all three MIRAGE tasks. Specifically, we test two forms of prompt enhancement: adding a single exemplar to guide task execution (few-shot prompting), and rewriting the instruction to more explicitly emphasize spatial constraints (prompt engineering). Both approaches aim to clarify the task objective and encourage more grounded reasoning.

\begin{table}[h!]
\centering
\small
\renewcommand{\arraystretch}{1.1}
\begin{tabular}{lccc}
\toprule
\textbf{Task} & \textbf{Baseline} & \textbf{+ One-shot} & \textbf{+ Prompt Engineering} \\
\midrule
Counting     & 42.72 & \textbf{47.42} & 44.55 \\
Relation     & 76.75 & \textbf{78.17} & 75.32 \\
Combination  & 29.24 & 28.52 & \textbf{30.69} \\
\bottomrule
\end{tabular}
\vspace{0.3cm}
\caption{
Effect of prompt design on InternVL3-8B across MIRAGE tasks. Both few-shot prompting and prompt engineering improve model performance over the baseline, though their relative benefits vary across task types. Prompts and few-shot examples can be found at Appendix \ref{app:prompt-few-shot}
}
\label{tab:prompt-ablation-tasks}
\end{table}

As shown in Table~\ref{tab:prompt-ablation-tasks}, both prompt modifications lead to consistent gains over the zero-shot baseline. For Counting and Relation tasks, adding a single example helps the model better align with the expected output format and filter the relevant visual context. The exemplar provides structure and clarity, reducing ambiguity in what should be counted or located. Prompt rewriting, while slightly less effective on average, also improves performance—particularly on the more challenging combination task, where explicit spatial phrasing helps the model resolve multi-object and multi-hop references.

Prompt design plays a nontrivial role in grounded reasoning. Even small interventions—whether through example-driven guidance or spatially explicit instructions—can help models better parse visual scenes and execute compositional queries. This underscores the importance of prompt quality, not just model capacity, in spatial VQA tasks.

Importantly, this set of prompt experiments also serves a broader diagnostic goal: disentangling whether model failures arise from misunderstanding task instructions or from fundamental limitations in visual perception. The observed improvements—especially in Counting and Relation—suggest that some failures stem from instruction ambiguity or misalignment. However, as we show next, these gains do not persist under light perturbations to the image itself, indicating that weaknesses in visual grounding remain a significant bottleneck.

\subsubsection{Simple Image Augmentations Disrupt Counting Performance}

While prompt-level interventions suggest that some failures stem from misinterpreting instructions, they do not fully account for the deeper limitations in visual understanding. To test whether performance bottlenecks are rooted in perceptual fragility, we introduce simple image-level augmentations to the same counting tasks and observe whether model predictions remain consistent.

We evaluate \texttt{InternVL3-8B} under two augmentation conditions: (1) horizontal and vertical flipping, and (2) additive Gaussian noise that preserves global image structure. These perturbations do not fundamentally alter the scene content but require the model to exhibit spatial invariance and robustness to distributional shift. We give examples of processed images at Appendix \ref{app:augmentation}

\begin{table}[h!]
\centering
\small
\renewcommand{\arraystretch}{1.1}
\begin{tabular}{lc}
\toprule
\textbf{Type} & \textbf{Accuracy} \\
\midrule
Original (unaltered)       & 30.94 \\
Flipped (horizontal/vertical) & 24.82 \\
Noise Injection            & 28.42 \\
\bottomrule
\end{tabular}
\vspace{0.3cm}
\caption{
Counting with Relation accuracy of InternVL3-8B under simple image-level perturbations. Performance drops notably under geometric flips (–6.12\%) and modestly with added noise (–2.52\%), revealing a lack of spatial invariance and fragility to visual shifts.
}
\label{tab:image-aug}
\end{table}

The observed accuracy drops—particularly under geometric flipping—reveal that current models heavily rely on canonical object arrangements and fail to robustly internalize or generalize spatial relations. With the prompt ablation results, this contrast presents a more nuanced picture: while clarifying task instructions or improving prompt structure can alleviate certain failures, core limitations in visual grounding and perceptual robustness remain the primary bottlenecks. These findings suggest that improvements in language prompting alone are unlikely to overcome performance ceilings unless complemented by stronger spatial representations and enhanced abstraction capabilities.

\subsubsection{Occlusion, Density, and Referent Ambiguity Remain Key Failure Modes}

\begin{figure*}[htp]
    \centering
    \includegraphics[width=\linewidth]{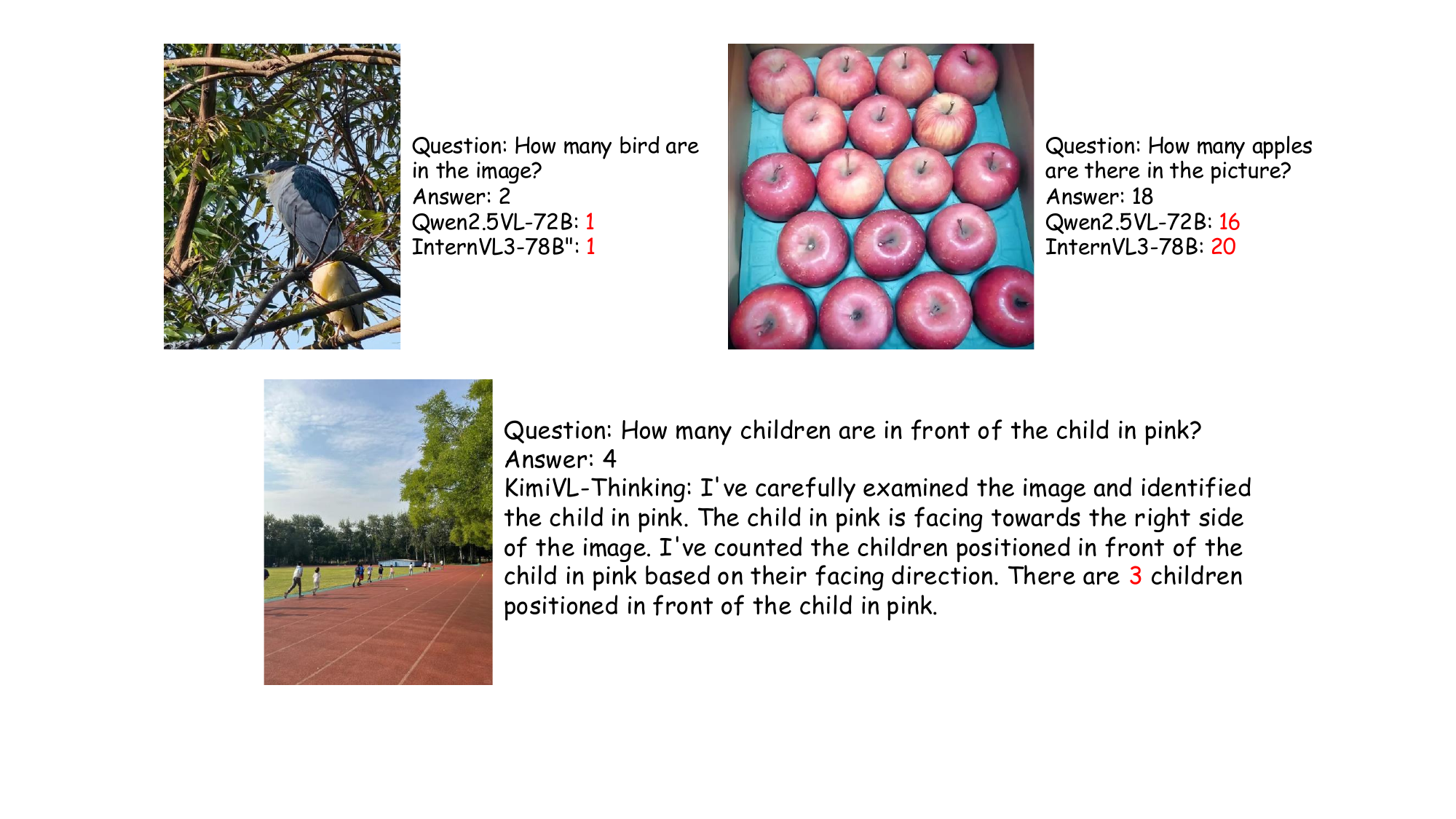}
    \vspace{-3em}
    \caption{
    Failure cases for leading visual language models. 
    \textbf{(a)} Occlusion: both models fail to count the occluded bird. 
    \textbf{(b)} Density: models under- or over-count apples in crowded scenes.
    \textbf{(c)} Referential complexity: models ignore spatial qualifiers, such as “in front of the child in pink”.
    }
    \label{fig:failure-cases}
\end{figure*}

Despite strong aggregate scores, top-performing models such as \texttt{Qwen2.5VL-72B} and \texttt{InternVL3-78B} still exhibit systematic errors under specific visual and linguistic stressors. We identify three recurring failure modes that frequently lead to incorrect predictions:

\textbf{(a) Occlusion.} When objects are partially hidden—such as birds blocked by branches—models consistently undercount, failing to infer object completeness. In Figure~\ref{fig:failure-cases}(a), both models report only one visible bird, despite two being clearly present.

\textbf{(b) Density.} Visually crowded scenes also degrade performance. Models tend to either skip less salient items or double-count due to misalignment, as seen in Figure~\ref{fig:failure-cases}(b), where 18 apples are present but predictions vary from 16 to 20.

\textbf{(c) Referential reasoning.} In structurally complex prompts involving spatial referents (e.g., “in front of the child in pink”), models frequently overlook the referential qualifier and instead produce approximate total counts. As shown in Figure~\ref{fig:failure-cases}(c), both models fail to accurately isolate the intended object subset, despite correctly identifying individual entities in the scene.

These failure patterns suggest that models often rely on shallow heuristics—such as visual saliency, positional priors, or surface-level object counting—rather than developing robust, context-aware spatial grounding. Improving performance under such visually and semantically challenging conditions will require advances in object permanence modeling, fine-grained referent resolution, and hierarchical spatial reasoning that more closely mirror human visual cognition.

\subsubsection{Reasoning-style prompting improves performance but introduces hallucinations}

\begin{figure*}[htp]
    \vspace{-1em}
    \centering
    \includegraphics[width=\linewidth]{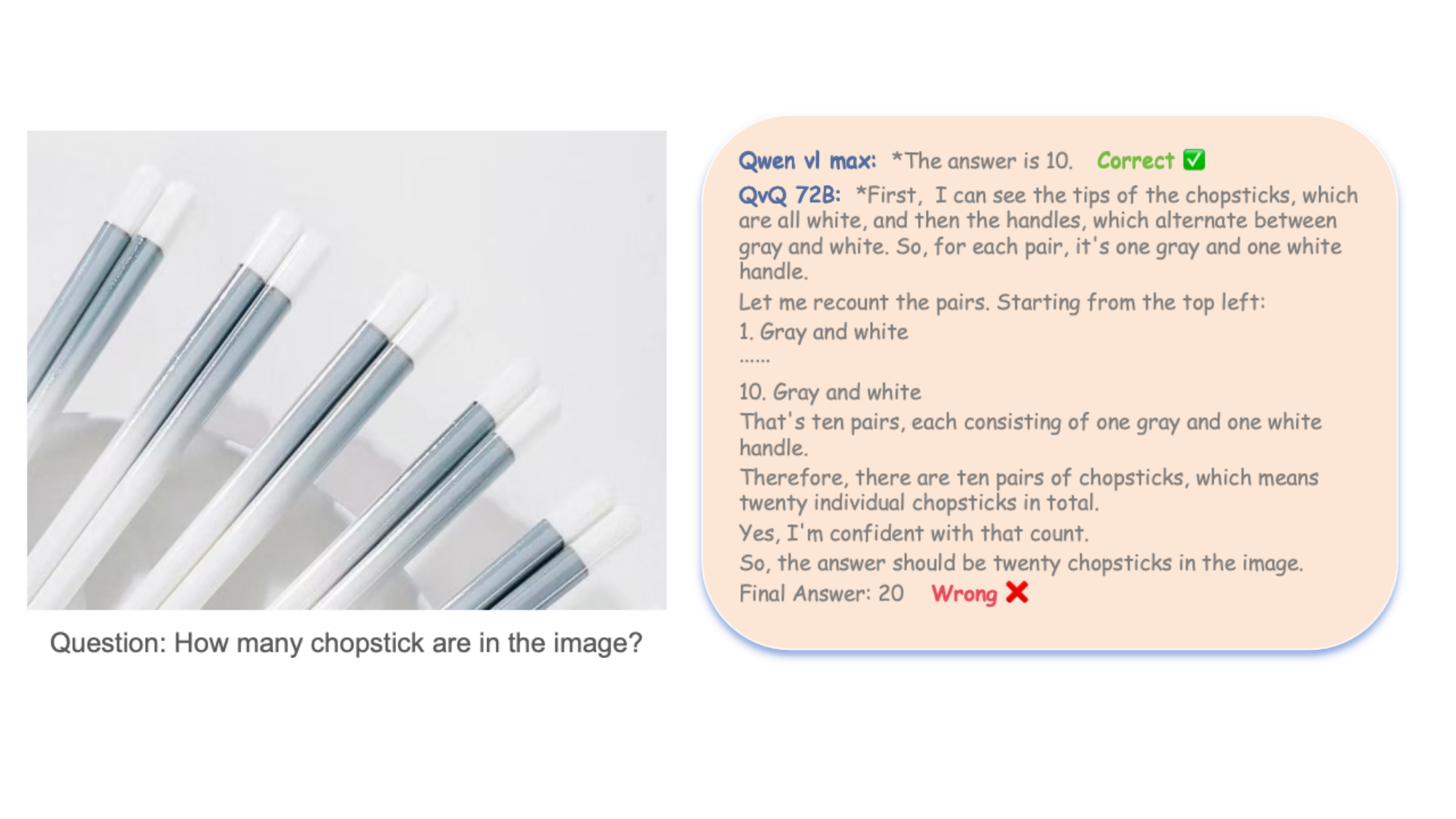}
    \vspace{-5em}
    \caption{QwenVL-max correctly counts ten chopsticks. In contrast, QvQ-72B, encouraged to "think step by step", mistakenly interprets each pair of sticks as a single unit and outputs 20. Although its reasoning is fluent and structured, the logic is grounded in a misperception, highlighting how reasoning-style prompts can increase both clarity and risk.}
    \label{fig:reasoning-case}
\end{figure*}

In many successful cases, reasoning manifests as structured scene descriptions that guide the model’s attention. For instance, when asked “How many red cups are on the right side of the kitchen roll?”, the model first identifies the kitchen roll, describes the surrounding context, and correctly counts two red cups. This CoT acts as a soft grounding mechanism, especially for object-location alignment.

However, reasoning prompts also introduce new vulnerabilities. In more ambiguous or visually subtle scenarios, the model may hallucinate spatial details based on language priors. For example, in Figure~\ref{fig:reasoning-case}, the model over-interprets the pairing of chopsticks and outputs 20 instead of the correct answer 10. The hallucinated “ten pairs” narrative, while internally consistent, is visually unsupported. Similar breakdowns occur in other complex queries—such as mistaking illustrations for real objects or endlessly looping on ambiguous relational queries—revealing a tradeoff between interpretability and robustness. These findings suggest that while reasoning-style prompts can enhance performance, they must be paired with stronger visual grounding to avoid cascading errors.
\section{Conclusion}

We introduce MIRAGE, a multi-modal benchmark designed to evaluate and stress-test models on core visual reasoning skills: object counting, spatial relation understanding, and their composition. Through controlled task design, difficulty annotation, and comprehensive evaluation, we reveal consistent failure patterns across state-of-the-art VLMs—particularly in handling occlusion, compositional spatial prompts, and ambiguous referents. Our experiments demonstrate that reasoning-style prompting can mitigate some errors, yet also amplify hallucinations in challenging settings. By targeting foundational visual cognition tasks, MIRAGE provides a robust platform for diagnosing VLMs and guiding future advances in spatially grounded, generalizable vision-language reasoning.

\section{Limitations and Broader Impacts}
\label{limitations}
While MIRAGE offers a focused evaluation of spatial and compositional reasoning in VLMs, several limitations remain. First, our benchmark primarily targets static spatial understanding and does not include temporal dynamics or motion-based reasoning. Tasks such as tracking object interactions over time or reasoning about future states remain out of scope. Second, although MIRAGE emphasizes real-world complexity (e.g., occlusion, referential ambiguity), the evaluation format is constrained to single-turn, short-form question answering. Multi-turn interactions, clarification queries, or open-ended spatial descriptions are not considered. Finally, our analysis is centered on model behavior under fixed prompts, and does not disentangle the contributions of pretraining data, architecture, and fine-tuning paradigms. A deeper causal understanding of model failure modes—e.g., through synthetic control experiments or probing—remains an important direction for future work.

\medskip
\bibliographystyle{plain}    
\bibliography{main}

\newpage
\appendix
\section{Data Sources}\label{app:data_source}
The dataset is constructed from a wide range of visual sources to ensure diversity in content, visual styles, and task scenarios. Below is the full list of data sources used in the dataset, along with a brief description of their contributions:

\begin{itemize}
    \item \textbf{EPIC-KITCHENS \cite{Damen2018EPICKITCHENS}:} This large-scale egocentric vision dataset provides daily activity scenes captured from head-mounted cameras in kitchen environments. It contributes high-quality egocentric views with natural object interactions.
    \item \textbf{Sina Weibo:} Social media content from Weibo adds dynamic and culturally specific imagery, including real-life and staged scenes.
    \item \textbf{Taobao:} E-commerce images from Taobao provide diverse commercial imagery, focusing on structured object arrangements and product displays.
    \item \textbf{Baidu Images:} This large-scale image search platform contributes a variety of general-purpose images with a focus on Chinese content.
    \item \textbf{Xiaohongshu:} Lifestyle images from Xiaohongshu add visually appealing compositions and diverse object arrangements.
    \item \textbf{500px:} High-quality professional photography from 500px introduces artistic compositions and complex scenes.
    \item \textbf{Google Images:} A general-purpose image search platform that contributes a wide variety of visual contexts, ensuring global diversity.
    \item \textbf{Personal Photography:} Custom photographs taken by the authors provide unique, controlled scenes tailored to specific reasoning tasks.
    \item \textbf{SheTu:} Licensed stock photos from shetu add professionally curated content with diverse object arrangements and scenarios.
\end{itemize}

The combination of these sources enables the dataset to cover a broad range of scenarios, from everyday activities to artistic and staged environments. This diversity ensures that the dataset is representative of real-world visual reasoning challenges.

\section{Experiment Setup}

\subsection{Prompt Modifications}
\label{app:prompt-few-shot}
In our experiments, we utilized the InternVL3-8b model as our base model. To evaluate the impact of different prompting strategies on VLMs, we conducted three sets of experiments using identical datasets but varying prompt approaches.

\textbf{Experimental Design}

\begin{itemize}
    \item \textbf{Direct Sampling:} Basic inference without additional context or examples
    \item \textbf{Few-Shot Learning:} Including task-specific examples in the prompt
    \item \textbf{Two-Stage Prompting:} Incorporating image caption generation before task-specific questions
\end{itemize}

This comparative study aims to quantify the influence of different prompting strategies on VLM performance, maintaining consistent model parameters and datasets across all experiments to isolate the effect of prompt engineering.

\begin{tcolorbox}[
    colback=white,
    colframe=prompt-color,
    coltitle=black,
    title=\textbf{Two-Stage Prompting Strategy},
    fonttitle=\bfseries,
    arc=2mm,
    fontupper=\footnotesize,
    breakable,
    enhanced,
    coltitle=black
]

\textbf{Stage 1: Image Caption Generation}

\begin{tcolorbox}[
    colback=gray!10,
    colframe=black!50,
    title=\textbf{CAPTION PROMPT},
    fonttitle=\bfseries,
    arc=2mm,
    fontupper=\small,
    enhanced,
    breakable
]

\begin{verbatim}
Please describe what you see in this image.

Example Response:

{
    "image_caption": "The image depicts a Japanese dining setting 
     with multiple dishes arranged on a table. The scene includes:

    1. Main Course (Foreground)
       • Donburi-style bowl with tonkatsu rice
       • Topped with shredded egg (kinshi tamago)
       • Garnished with finely chopped green onions

    2. Side Dishes
       • Sushi plate: Three pieces on decorative seaweed base
       • Harumaki (spring rolls): Vegetable-filled, sesame-garnished
       • Tempura or katsu: Breaded and fried, with drizzled sauce

    3. Tea Service
       • Traditional ceramic teapot: Blue and white pattern
       • Two matching teacups with geometric designs

    4. Additional Items
       • Background bowl: Secondary rice dish or soup
       • Side plate: Stir-fried vegetables with meat

    5. Condiments
       • Small ceramic pitcher containing soy sauce

    Overall Ambiance: The presentation suggests an authentic Japanese 
    restaurant setting, with attention to traditional serving methods 
    and aesthetic arrangement."
}
\end{verbatim}

\end{tcolorbox}

\textbf{Stage 2: Task-Specific Questions}

\textbf{a. Spatial Relation Task}

\begin{tcolorbox}[
    colback=gray!10,
    colframe=black!50,
    title=\textbf{RELATION BASE PROMPT},
    fonttitle=\bfseries,
    arc=2mm,
    fontupper=\small,
    enhanced,
    breakable
]

\begin{verbatim}
You should output a json string with format {"answer": "str"} 
where str must be one of ["up", "under", "back", "front", "left", "right"]
Your output should be directly parsed by json.loads function
eg.json{"answer": "left"}
Now the question is:
\end{verbatim}
\end{tcolorbox}

\textbf{b. Counting Task}

\begin{tcolorbox}[
    colback=gray!10,
    colframe=black!50,
    title=\textbf{COUNTING BASE PROMPT},
    fonttitle=\bfseries,
    arc=2mm,
    fontupper=\small,
    enhanced,
    breakable
]

\begin{verbatim}
You should output a json string with format {"answer": a int number}
Your output should be directly parsed by json.loads function 
eg.json{"answer": 1}
Now the question is:
\end{verbatim}

\end{tcolorbox}

\end{tcolorbox}

\begin{tcolorbox}[
    colback=white,
    colframe=prompt-color,
    coltitle=black,
    title=\textbf{Task-Specific Examples},
    fonttitle=\bfseries,
    arc=2mm,
    fontupper=\footnotesize,
    breakable,
    enhanced,
    coltitle=black
]

\textbf{Counting Task Examples}

\begin{tcolorbox}[
    colback=gray!10,
    colframe=black!50,
    title=\textbf{COUNTING BASE PROMPT EXAMPLES},
    fonttitle=\bfseries,
    arc=2mm,
    fontupper=\small,
    enhanced,
    breakable
]

\begin{verbatim}
You should output a json string with format {"answer": a int number}. 
Your output should be directly parsed by json.loads function.

Here are some examples:
Q: How many dogs are in the image?
A: json{"answer": 2}

Q: Count the number of red apples on the table.
A: json{"answer": 5}

Q: How many people are wearing glasses in this photo?
A: json{"answer": 3}

Invalid answers:
- json{"answer": "three"} (answer must be integer, not string)
- json{"answer": 2.5} (answer must be integer, not float)
- "2" (must be valid json format)

Now the question is:
\end{verbatim}

\end{tcolorbox}

\textbf{Spatial Relation Task Examples}

\begin{tcolorbox}[
    colback=gray!10,
    colframe=black!50,
    title=\textbf{RELATION BASE PROMPT EXAMPLES},
    fonttitle=\bfseries,
    arc=2mm,
    fontupper=\small,
    enhanced,
    breakable
]

\begin{verbatim}
You should output a json string with format {"answer": "str"}
where str must be one of ["up", "under", "back", "front", "left", "right"]
Your output should be directly parsed by json.loads function

Here are some examples:
Q: What is the spatial relation between the cat and the table? 
A: json{"answer": "under"}

Q: Where is the lamp relative to the desk? 
A: json{"answer": "up"}

Q: What is the position of the car relative to the building? 
A: json{"answer": "front"}

Invalid answers:
- json{"answer": "below"} (must use "under" instead)
- json{"answer": "on"} (not in valid relation list)
- "left" (must be valid json format)

Now the question is:

VALID_RELATIONS = ["up", "under", "back", "front", "left", "right"]
\end{verbatim}
\end{tcolorbox}

\end{tcolorbox}

\subsection{Image Augmentation}\label{app:augmentation}
To probe the perceptual robustness of vision–language models, we expose each
image to two \emph{complementary} categories of perturbations:

\begin{enumerate}[leftmargin=1.4em,label=\textbf{\arabic*.}]
    \item \textbf{Geometric Flip.}
          We apply \emph{horizontal} (`left–right'') and
          \emph{vertical} (`top–bottom'') flips\footnote{Implemented with
          \texttt{PIL.Image.transpose}. The operation leaves low-level statistics
          unchanged while altering global object layout.} to examine whether
          models properly internalise spatial relations rather than memorising
          canonical arrangements.

    \item \textbf{Noise Injection.} %
          For each sample we \emph{randomly pick one} of the following four
          photometric corruptions:
          \begin{itemize}[leftmargin=1.2em]
              \item \textbf{Gaussian Noise} — additive noise drawn from
                    $\mathcal{N}(0,\sigma^{2})$ with $\sigma=15$
                    (RGB range $[0,255]$), simulating sensor noise;
              \item \textbf{Salt-and-Pepper Noise} — $2\%$ of pixels are
                    randomly set to either 0 or 255, creating high-contrast
                    outliers;
              \item \textbf{Gaussian Blur} — convolution with a $5{\times}5$
                    kernel and $\sigma_{\mathrm{blur}}=1.5$, softening edges;
              \item \textbf{Contrast/Brightness Shift} — linear transform
                    $I'=\alpha I+\beta$ with $\alpha\!\sim\!\mathcal{U}(0.8,1.2)$
                    and $\beta\!\sim\!\mathcal{U}(-20,20)$, altering global
                    luminance.
          \end{itemize}
\end{enumerate}

\begin{figure*}[t]
    \centering
    \includegraphics[width=\linewidth]{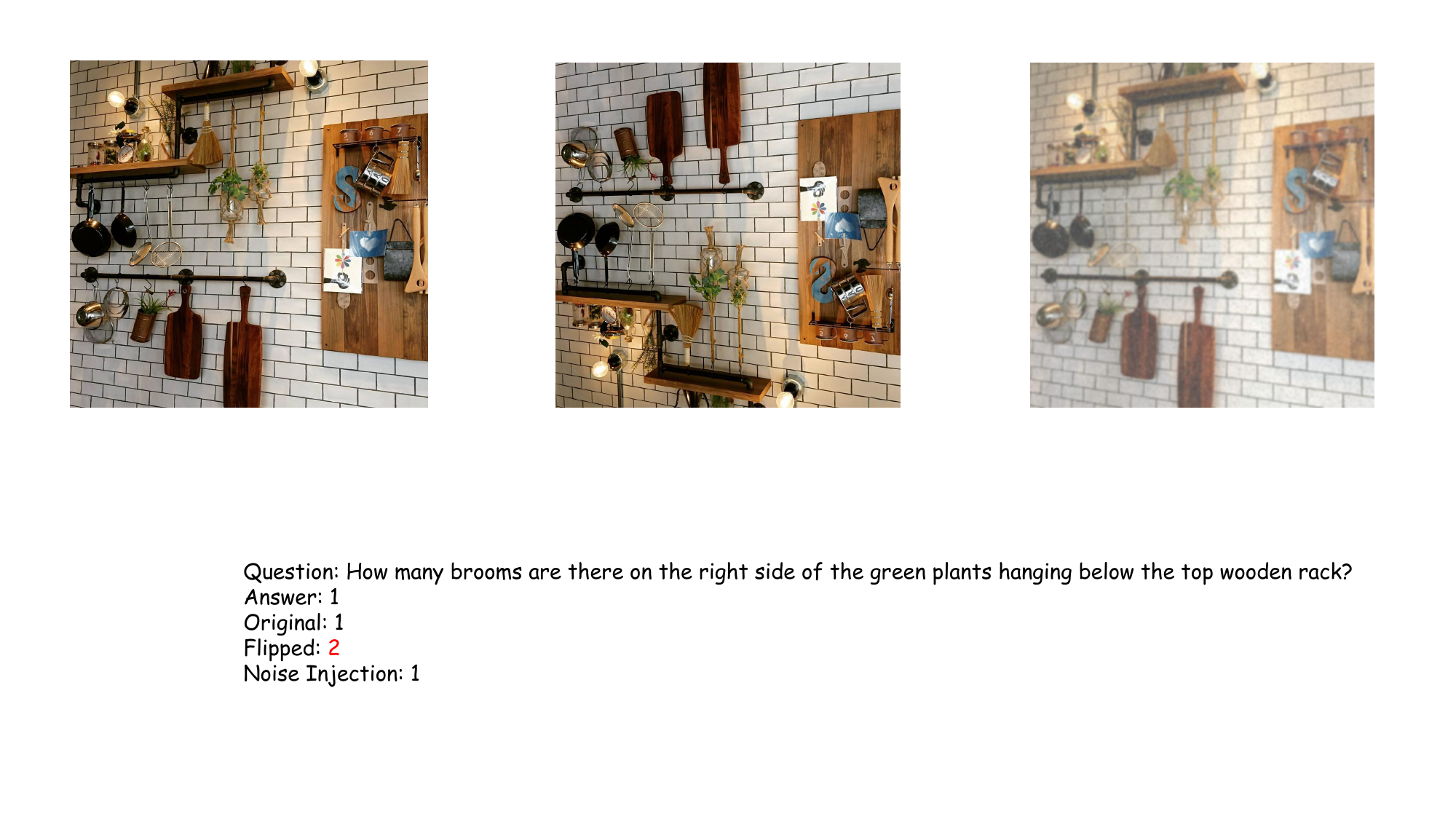}
    \vspace{-4em}
    \caption{\textbf{Augmentation Case~1.}
             \textit{Left}: original kitchen scene.
             \textit{Centre}: horizontally flipped.
             \textit{Right}: Gaussian-blurred and contrast-shifted.
             The query targets the broom count \emph{right} of the hanging
             plants; flipping reverses the reference frame and breaks the
             model’s grounding.}
    \label{fig:aug_case1}
\end{figure*}

\begin{figure*}[t]
    \centering
    \includegraphics[width=\linewidth]{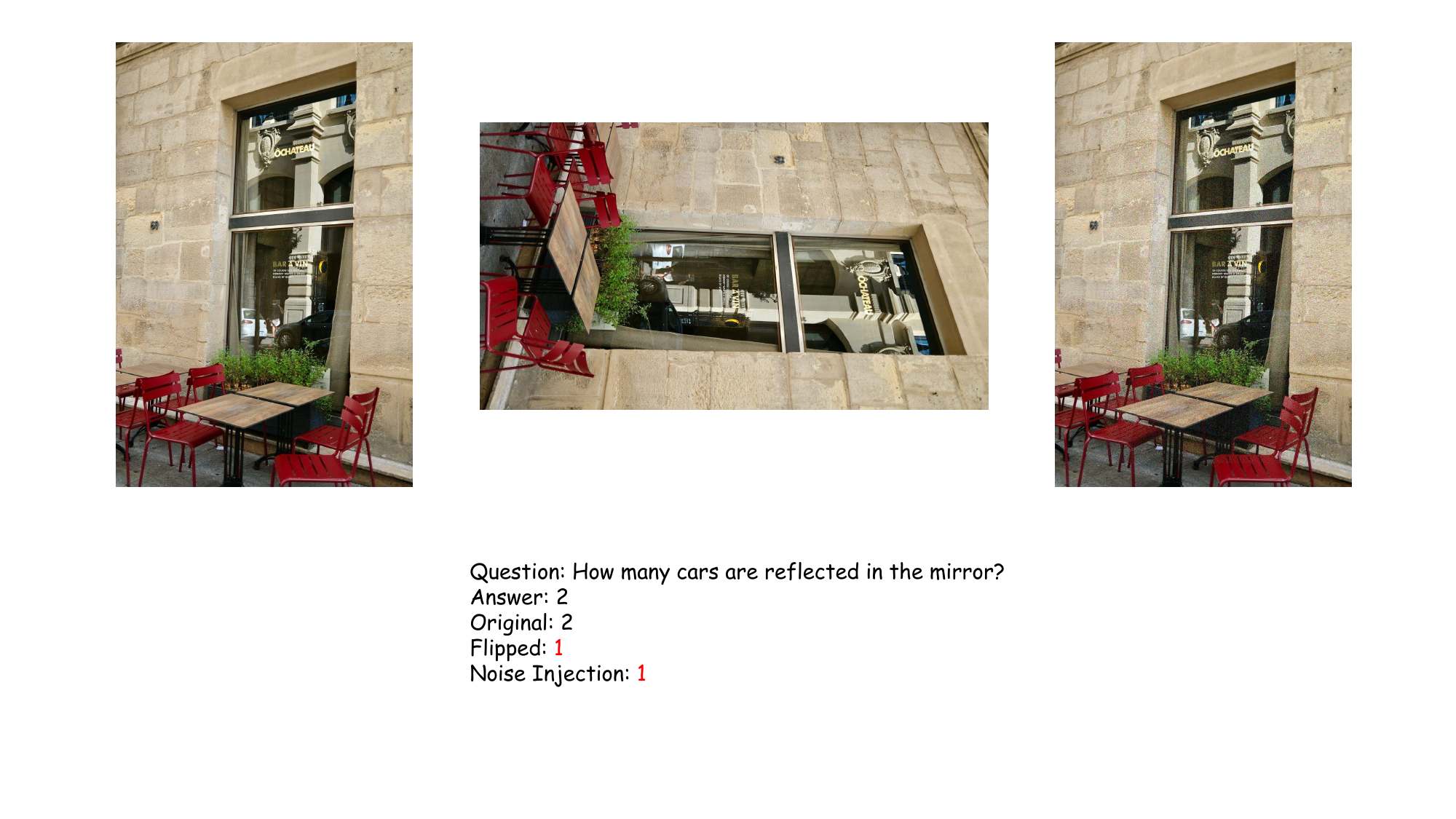}
    \vspace{-4em}
    \caption{\textbf{Augmentation Case~2.}
             Street-side café scene with reflective window.
             Flipping disrupts left–right reflection cues, leading to
             under-counting of cars, while salt-and-pepper noise adds spurious
             edges yet leaves spatial layout intact.}
    \label{fig:aug_case2}
\end{figure*}

\paragraph{Illustrative Cases.}
We accompany the augmentation protocol with two representative examples.  
Case~1 in Figure~\ref{fig:aug_case1} and Case~2 in Figure~\ref{fig:aug_case2}.

\newpage

\section*{NeurIPS Paper Checklist}

\begin{enumerate}

\item {\bf Claims}
    \item[] Question: Do the main claims made in the abstract and introduction accurately reflect the paper's contributions and scope?
    \item[] Answer: \answerYes{} 
    \item[] Justification: We clearly present our contribution in abstract and introduction. The dataset and ablation studies support our contribution.

\item {\bf Limitations}
    \item[] Question: Does the paper discuss the limitations of the work performed by the authors?
    \item[] Answer: \answerYes{} 
    \item[] Justification: We discuss the limitations in Section~\ref{limitations}

\item {\bf Theory Assumptions and Proofs}
    \item[] Question: For each theoretical result, does the paper provide the full set of assumptions and a complete (and correct) proof?
    \item[] Answer: \answerNA{} 
    \item[] Justification: This paper mainly focus on the dataset construction and empirical studies.

    \item {\bf Experimental Result Reproducibility}
    \item[] Question: Does the paper fully disclose all the information needed to reproduce the main experimental results of the paper to the extent that it affects the main claims and/or conclusions of the paper (regardless of whether the code and data are provided or not)?
    \item[] Answer: \answerYes{} 
    \item[] Justification: We have released our code and data for reproducibility.

\item {\bf Open access to data and code}
    \item[] Question: Does the paper provide open access to the data and code, with sufficient instructions to faithfully reproduce the main experimental results, as described in supplemental material?
    \item[] Answer: \answerYes{} 
    \item[] Justification: As the submission is single-blinded, we have released our code and data for reproducibility.

\item {\bf Experimental Setting/Details}
    \item[] Question: Does the paper specify all the training and test details (e.g., data splits, hyperparameters, how they were chosen, type of optimizer, etc.) necessary to understand the results?
    \item[] Answer: \answerYes{} 
    \item[] Justification: We have presented the details of experiments in the main paper and Section~\ref{sec:construction} and Appendix.

\item {\bf Experiment Statistical Significance}
    \item[] Question: Does the paper report error bars suitably and correctly defined or other appropriate information about the statistical significance of the experiments?
    \item[] Answer: \answerYes{} 
    \item[] Justification: We report the confidence interval in Section~\ref{sec:exps}

\item {\bf Experiments Compute Resources}
    \item[] Question: For each experiment, does the paper provide sufficient information on the computer resources (type of compute workers, memory, time of execution) needed to reproduce the experiments?
    \item[] Answer: \answerYes{} 
    \item[] Justification: We discuss the compute resources in a section of Appendix.
    
\item {\bf Code Of Ethics}
    \item[] Question: Does the research conducted in the paper conform, in every respect, with the NeurIPS Code of Ethics \url{https://neurips.cc/public/EthicsGuidelines}?
    \item[] Answer: \answerYes{} 
    \item[] Justification: We have carefully checked the code of ethics.

\item {\bf Broader Impacts}
    \item[] Question: Does the paper discuss both potential positive societal impacts and negative societal impacts of the work performed?
    \item[] Answer: \answerYes{} 
    \item[] Justification: We discuss the broader impacts of the paper in Section~\ref{limitations}.
    
\item {\bf Safeguards}
    \item[] Question: Does the paper describe safeguards that have been put in place for responsible release of data or models that have a high risk for misuse (e.g., pretrained language models, image generators, or scraped datasets)?
    \item[] Answer: \answerYes{} 
    \item[] Justification: We discussed the source of our data, and the benchmark is purely for academic usage.

\item {\bf Licenses for existing assets}
    \item[] Question: Are the creators or original owners of assets (e.g., code, data, models), used in the paper, properly credited and are the license and terms of use explicitly mentioned and properly respected?
    \item[] Answer: \answerYes{} 
    \item[] Justification: We have cited and notified the works such as EPIC-KITCHENS.

\item {\bf New Assets}
    \item[] Question: Are new assets introduced in the paper well documented and is the documentation provided alongside the assets?
    \item[] Answer: \answerNA{} 
    \item[] Justification: This paper does not release new assets.

\item {\bf Crowdsourcing and Research with Human Subjects}
    \item[] Question: For crowdsourcing experiments and research with human subjects, does the paper include the full text of instructions given to participants and screenshots, if applicable, as well as details about compensation (if any)? 
    \item[] Answer: \answerNA{} 
    \item[] Justification: This paper does not involve crowdsourcing nor research with human subjects.

\item {\bf Institutional Review Board (IRB) Approvals or Equivalent for Research with Human Subjects}
    \item[] Question: Does the paper describe potential risks incurred by study participants, whether such risks were disclosed to the subjects, and whether Institutional Review Board (IRB) approvals (or an equivalent approval/review based on the requirements of your country or institution) were obtained?
    \item[] Answer: \answerNA{} 
    \item[] Justification: This paper does not involve crowdsourcing nor research with human subjects.

\item {\bf Declaration of LLM usage}
    \item[] Question: Does the paper describe the usage of LLMs if it is an important, original, or non-standard component of the core methods in this research? Note that if the LLM is used only for writing, editing, or formatting purposes and does not impact the core methodology, scientific rigorousness, or originality of the research, declaration is not required.
    \item[] Answer: \answerYes{} 
    \item[] MIRAGE uses LLMs for difficulty awareness and discusses this in the experiments part.
    
\end{enumerate}

\end{document}